**Data distribution impacts the performance and generalisability of contrastive learning-based foundation models of electrocardiograms**

Gul Rukh Khattak[1], Konstantinos Patlatzoglou[1], Joseph Barker[1], Libor Pastika[1], Boroumand Zeidaabadi[1], Ahmed El-Medany[1,2], Hesham Aggour[1], Yixiu Liang[1,3], Antonio H. Ribeiro [4], Jeffrey Annis[5,6,7], Antonio Luiz Pinho Ribeiro[8], Junbo Ge[2], Daniel B. Kramer[9], Jonathan W. Waks[10], Evan Brittain[5,6], Nicholas Peters[1], Fu Siong Ng*[1,11], Arunashis Sau*[1,11]

*joint senior authors

**Affiliations**

1. National Heart and Lung Institute, Imperial College London, United Kingdom.
2. Department of Cardiology, Chelsea and Westminster Hospital NHS Foundation Trust, London, United Kingdom
3. Department of Cardiology, Zhongshan Hospital of Fudan University, Shanghai Institute of Cardiovascular Diseases, National Clinical Research Centre for Interventional Medicine, Shanghai, China
4. Department of Information Technology, Uppsala University, Sweden.
5. Vanderbilt Institute for Clinical and Translational Research, Vanderbilt University Medical Center, Nashville, TN, USA.
6. Center for Digital Genomic Medicine, Department of Medicine, Vanderbilt University Medical Center, Nashville, TN, USA.
7. Division of Cardiovascular Medicine, Vanderbilt University Medical Center, Nashville, TN, USA

8    Department of Internal Medicine, Faculdade de Medicina, and Telehealth Center and Cardiology Service, Hospital das Clínicas, Universidade Federal de Minas Gerais, Belo Horizonte, Brazil

9    Richard A. and Susan F. Smith Center for Outcomes Research in Cardiology, Beth Israel Deaconess Medical Center, Harvard Medical School, Boston MA USA

10   Harvard-Thorndike Electrophysiology Institute, Beth Israel Deaconess Medical Center, Harvard Medical School, Boston, MA, USA.

11   Department of Cardiology, Imperial College Healthcare NHS Trust, London, United Kingdom

**Disclosures:**

JWW and DBK were previously on the advisory board for Heartcor solutions LLC, for whom they remain independent consultants. JWW reports research funding from Anumana and is a consultant for HeartBeam Inc. FSN reports speaker fees from GE healthcare and is on the advisory board for Astra Zeneca. AS, LP, BZ and FSN declare inventorship on a patent application relating to AI-ECG methods. The remaining authors have no conflicts to declare.

**Correspondence:**

Gul Rukh Khattak

Post-doctoral Research Associate

*e-mail: g.khattak@imperial.ac.uk*

and

Arunashis Sau

Academic Clinical Lecturer

*e-mail: as7909@imperial.ac.uk*

and

Fu Siong Ng

Professor of Cardiac Electrophysiology

*e-mail: f.ng@imperial.ac.uk*

National Heart and Lung Institute, Imperial College London

Hammersmith Campus

Du Cane Road

London W12 0NN

**Abstract**

Contrastive learning is a widely adopted self-supervised pretraining strategy, yet its dependence on cohort composition remains underexplored. We present Contrasting by Patient Augmented Electrocardiograms (**CAPE**) foundation model and pretrain on four cohorts (n = 5,203,352), from diverse populations across three continents (North America, South America, Asia). We systematically assess how cohort demographics, health status, and population diversity influence the downstream performance for prediction tasks also including two additional cohorts from another continent (Europe). We find that downstream performance depends on the distributional properties of the pretraining cohort, including demographics and health status. Moreover, while pretraining with a multi-centre, demographically diverse cohort improves in-distribution accuracy, it reduces out-of-distribution (OOD) generalisation of our contrastive approach by encoding cohort-specific artifacts. To address this, we propose the **In-Distribution Batch (IDB)** strategy, which preserves intra-cohort consistency during pretraining and enhances OOD robustness. This work provides important insights for developing clinically fair and generalisable foundation models.

**Introduction**

Electrocardiograms (ECGs) provide a non-invasive and widely accessible means of recording the heart's electrical activity, capturing myocardial depolarisation and repolarization across multiple leads positioned on the skin. Despite the potential to reveal important physiological insights, the clinical utility of ECGs remains in part constrained by the need for expert interpretation. Additionally, recent studies have demonstrated the potential for artificial intelligence-enhanced ECG (AI-ECG) analysis to identify undiagnosed disease and predict risk of adverse events better than human experts [1,2,3,4].

Pretraining enables AI models to learn general patterns from large datasets, which enhances performance, accelerates learning, and reduces the need for extensive task-specific data. The abundance of unlabeled ECG data, coupled with the high cost of expert annotation, highlights the promise of self-supervised learning (SSL) as an effective pretraining strategy for ECGs. Among SSL approaches, contrastive learning has gained prominence in the medical domain[5], for its ability to learn meaningful representation (features) by promoting similarity between contextually positive pairs and dissimilarity from negatives[6]. Contrastive learning has been applied to ECGs with various definitions of positive and negative pairs involving contrasting by different signals[7-9] or applying a range of signal transformations (augmentations) to the same signal[10-12]. The effectiveness of a pretraining approach is influenced by the characteristics of the dataset used. In image-based models, studies have shown that non-curated datasets can yield stronger representation learning than carefully curated ones[13]. However, a similarly large-scale, systematic investigation of pretraining datasets has yet to be conducted for electrocardiogram (ECG) data. A previous study examined out-of-distribution (OOD) performance in ECG-based self-supervised pretraining by comparing supervised models trained on features from SSL-pretrained encoders across two medium size secondary cohorts cohorts, demonstrating comparable performance despite pronounced distributional shifts between them[14].

Deep learning models, often struggle with generalisation, particularly when evaluated on out-of-distribution (OOD) data[15, 16]. This is complicated further by the ability to learn complex patterns beyond human capabilities, as demonstrated by their ability to predict patient age or sex from ECG data[4]. However, this strength also makes them prone to overfitting the training data. Performance has been shown to vary with patient health status[17] and may be influenced by demographic factors such as ethnicity, which have been associated with identifiable signatures in ECG signals[18]. This challenge is further compounded by significant heterogeneity across ECG cohorts, stemming from differences in age distribution, clinical profiles, recording devices, and institutional practices. The ECG-based models experience degraded performance when tested on data from different sources, proving that domain shifts across the ECG cohorts can affect model generalisation[19].

Foundation models are pretrained on large-scale datasets to learn generalisable representations, and they must overcome domain shift to be clinically viable. High performance on in-distribution (ID) data is insufficient if the model fails to generalise to external, out-of-distribution (OOD) datasets. For example, a retinal image–based foundational model trained on British patient data demonstrated poor transferability to Chinese cohorts, performing comparably to model pretrained on natural images[20]. This highlights the critical need for robust generalisation across diverse populations to ensure the practical utility of ECG foundation models. Evidence of socio-demographic bias has been reported in foundation models used for medical diagnosis based on large language models[21].

Contrastive learning performance inherently depends on how positive and negative samples are defined, so we hypothesize that cohort distribution can significantly influence the quality of the learned features. We explore the generalisation and robustness of our **Contrasting by Augmented Patient Electrocardiograms (CAPE)** foundation model[22], which integrates within-patient contrastive strategy with signal-preserving augmentations such as random cropping and zero masking. Using a multi-centre cohort of over five million ECGs, we investigate how data diversity, label distribution, and cohort composition during pretraining affect downstream performance.

In this study, we investigate how cohort composition, clinical setting, and data diversity during pretraining influence the robustness and generalisability of ECG representation models. To address the challenges introduced by heterogeneous data sources, we further propose the **In-Distribution Batch (IDB)** approach, which constrains contrastive learning batches to single cohorts to reduce distributional noise. This work aims to guide the development of clinically meaningful foundation models that can generalise effectively across diverse patient populations and healthcare environments.

## Results

### Cohorts

This study utilizes a diverse set of large-scale electrocardiogram (ECG) cohorts from five countries across four continents: the Beth Israel Deaconess Medical Centre (BIDMC)[3] and Vanderbilt University Medical Centre (VUMC)[23] from the United States, the Clinical Outcomes in Digital Electrocardiography (CODE) cohort[24] from Brazil, the Shanghai Zhongshan Hospital (SHZS) cohort from China, the UK Biobank (UKB)[25] from the United Kingdom, and the Physikalisch-Technische Bundesanstalt (PTB-XL) dataset[26] from Germany. Among these, BIDMC, PTB-XL, and VUMC originate from secondary care settings and are characterized by a higher prevalence of cardiovascular disease.

Table 1 summarises the characteristics of the unlabelled data from patients with multiple ECGs used for contrastive pretraining from BIDMC, CODE, SHZS, and VUMC cohorts. The combined BCSV cohort includes over five million unlabelled ECGs drawn from BIDMC (127,041 patients; 1,106,886 ECGs), CODE (424,577 patients; 1,123,903 ECGs), SHZS (420,956 patients; 1,560,551 ECGs), and VUMC (252,306 patients; 1,412,012 ECGs). Mean patient ages are 57.99 years (BIDMC), 56.00 years (CODE),

52.08 years (SHZS), and 58.05 (VUMC); the female-to-male ratios are 50% (BIDMC), 61% (CODE), 44% (SHZS), and 49% (VUMC).

The downstream prediction tasks employ BIDMC, CODE, SHZS and together with additional cohorts including UKB, a large volunteer cohort comprising 70,655 ECGs from 66,402 patients (mean age 65.35 years; 52% female), and PTB-XL, an open-access secondary care dataset containing 21,799 ECGs from 18,869 patients (mean age 62.36 years; 51% female). The demographic and health characteristics of the cohorts used for label prediction are presented (Table 2).

**Pretraining**

The training pipeline consists of two stages: contrastive pretraining followed by supervised learning for prediction tasks. Initially, CAPE models are pretrained to learn generic ECG features, which are then generated for each labeled cohort. This approach significantly reduces training time for downstream tasks by leveraging these precomputed features. Supervised learning is performed using a multi-layer perceptron (MLP) prediction head. Age and sex serve as target labels, representing benchmarks for regression (age) and classification (sex) tasks, respectively. These labels are selected because they are available across all cohorts. Model performance is evaluated using mean absolute error (MAE) for age regression and area under the receiver operating characteristic curve (AUROC) for sex classification, consistent with prior work[17, 26].

**Composition of pretraining cohort affects performance**

We explore how the demographic composition of pretraining datasets influences the quality of learned ECG representational features and the accuracy of downstream predictive models. Five CAPE models pretrain on distinct cohorts with varied characteristics: BIDMC (CAPE-B), CODE (CAPE-C), SHZS (CAPE-S), VUMC (CAPE-V), and the multicentre BCSV dataset (CAPE-X) (Fig. 1). Each model extracts features from all label cohorts, which are then used to train multi-layer perceptron networks (MLP) for age and sex prediction, yielding twenty-five models per task. A consistent split of 10,000 training, 2,000 validation, and 2,000 test samples applies across all cohorts.

Performance is reported as mean absolute error (MAE) for age regression (Fig. 2a) and area under the ROC curve (AUROC) for sex classification (Fig. 2b), averaged over six runs. Models pretrained on the multi-centre BCSV dataset (CAPE-X) achieve the best overall performance, demonstrating the benefit of pretraining on a large and diverse dataset. CAPE-X achieves the lowest mean MAE for age prediction (7.86, 95% CI: 7.70–7.82, p=0.0002) and the highest mean AUROC for sex prediction (0.955, 95% CI: 0.955–0.956, p<0.0001). Among CAPE models pretrained on single cohorts, those trained on secondary care datasets such as BIDMC (age MAE: 8.19, 95% CI 8.10–8.28; sex AUROC: 0.942, 95% CI: 0.941–0.942) and VUMC (age MAE: 8.20, 95% CI: 8.12–08.27; sex AUROC: 0.942, 95% CI: 0.942–0.943) outperform those trained on healthier population-based cohorts such as CODE (age MAE: 8.40, 95% CI: 8.35–8.45; sex AUROC: 0.934, 95% CI: 0.934–0.935) and SHZS (age MAE:8.23, 95% CI: 8.16–8.30; sex AUROC: 0.924, 95% CI: 0.924–0.925, with differences significant at $p < 0.001$.

**Out-of-Distribution Performance Decline in Multi-Cohort Contrastive Pretraining**

We evaluate the out-of-distribution (OOD) generalisation capabilities of supervised prediction models trained on representations extracted from a single cohort and tested on independent external cohorts (Fig. 3). While prior analyses confirmed that CAPE-X achieves the highest average performance when models are both trained and tested on the same dataset, here we explore the robustness of these representations under the distributional shift by conducting supervised training on one cohort and testing on another. The CAPE models are pretrained on the BCSV dataset, with the BIDMC test split deliberately excluded to prevent information leakage and ensure an unbiased evaluation.

The CAPE-X model is pretrained on the combined BCSV dataset, with batches randomly sampled across multiple source cohorts. Following pretraining, feature representations are extracted for the BIDMC cohort (n = 1,169,387), and a separate multilayer perceptron classifier, designated MLP-XB, is trained for each prediction task. The BIDMC features are partitioned into training, validation, and test sets comprising 50 percent, 10 percent, and 40 percent of the data, respectively. Supervised training is repeated across ten independent runs using these predefined splits to ensure robust performance estimates. To assess OOD generalisation, each trained MLP-XB model is evaluated on randomly sampled subsets of 10,000 instances from each of the external cohorts using the same seed. This sampling strategy enables assessment of performance variability across independent runs while accounting for differences in cohort sizes and distributions, which is essential given the heterogeneity in cohort size and demographic composition.

Quantitative evaluation (Table 3) reveals a marked performance degradation of the MLP-XB for CAPE-X features when applied to external cohorts, particularly CODE (age MAE : 24.00, 95% CI 18.45–29.55 & Sex AUROC: 0.598, 95% CI 0.567–0.630) and SHZS (age MAE : 14.47, 95% CI 10.63–18.32 & Sex AUROC: 0.652, 95% CI 0.564–0.740), despite the broad and diverse pretraining base of CAPE-X.

Pretraining with In-Distribution Batches improves generalisation

To further investigate this degradation, we conducted a qualitative analysis using t-distributed stochastic neighbour embedding (t-SNE), a nonlinear dimensionality reduction technique for visualizing high-dimensional representations[27] (Fig. 4). The t-SNE projections of CAPE-X features (Fig. 4B) exhibit distinct clustering by cohort, indicating that the model encodes cohort-specific information. Additional visualizations of ethnicity distribution within BIDMC (Fig. 4B) and device types in CODE (Fig. 4C) suggest that these clusters are primarily driven by differences in acquisition hardware rather than patient demographics. By contrast, features generated by the CAPE-B model (described in previous section) display a continuous and more homogeneous distribution across cohorts (Fig. 4A), indicating a partial reduction in cohort specificity. These findings suggest that even with standardized preprocessing, technical variation introduced by different acquisition devices or pipeline artifacts can bias the learned representations and reduce their ability to generalise across settings.

To address this challenge, we introduce a novel training approach termed In Distribution Batching (IDB), implemented in the CAPE-Z model (Fig. 3). In this strategy, each training batch is composed exclusively of samples from a single cohort, ensuring that contrastive negative pairs originate from the same distribution. This design discourages the model from learning spurious cohort-distinguishing features and instead promotes the identification of clinically meaningful variability within each cohort. Features produced by CAPE-Z (Fig. 4D) show no discernible clustering or cohort-related gradients in t-SNE space, suggesting improved robustness to distributional differences. Consistent with these observations, In the CODE cohort, CAPE-Z achieves an age prediction MAE of 7.93 (95% CI: 7.60–8.26) and a sex classification AUROC of 0.927 (95% CI: 0.913–0.941), both statistically significant compared to CAPE-X ($p < 0.001$). Similarly, in the SHZS cohort, the age MAE is reduced to 7.24 (95% CI: 7.04–7.44), and the AUROC for sex classification reaches 0.935 (95% CI: 0.924–0.946), with highly significant improvements over CAPE-X ($p < 0.001$). These results highlight CAPE-Z's capacity to generalise effectively across diverse populations and previously challenging settings. These results highlight the efficacy of the IDB strategy in producing generalised ECG representations and support its potential to improve model reliability and applicability in diverse clinical environments.

**Discussion**

To our knowledge, this is the first systematic exploration of how different pretraining cohorts influence learned feature representations, and how varying downstream cohorts affect predictive performance on target labels. Within our contrastive learning framework, we find that combining multiple pretraining cohorts markedly reduces the generalisation ability of downstream models to out-of-distribution (OOD) data. To mitigate this effect, we introduce an In-Distribution Batch (IDB) strategy, which explicitly aligns pretraining distributions with the intended downstream application.

Foundation models leverage large-scale cohorts to learn rich and generalisable feature representations. This capability is particularly valuable in medical applications, where annotated datasets are typically limited in size and exhibit significant class imbalance or skewed distributions. The role of foundation models in medicine is rapidly expanding[28], demonstrating enhanced performance in diagnostic tasks across modalities such as X-rays[29], CT scans[30], and electrocardiograms[11]. Despite these advances, a fundamental challenge remains achieving robust generalisation to real-world clinical data, which frequently exhibit substantial distributional shifts[31]. Models trained on medical data from one cohort could not be generalised to another population from another continent across diverse modalities[18, 19, 20, 21].

**Pretraining data distribution affects the quality of the learned representations**

The influence of health status and demographic factors on learned representations is of critical importance for successful clinical translation. Within supervised learning paradigms, model performance has been observed to differ significantly between healthy and unhealthy populations, with models typically yielding superior results for healthy subjects[16]. Prior work has demonstrated the demographic sensitivity of foundation models, particularly in vision-language tasks involving chest X-ray classification[32], as well as in large language models applied to medical decision-making[21]. A novel analysis of the data distributions on three popular ECG-based arrhythmia datasets, namely PTB-XL (n=21,837), Chapman (n=10,646), and Ribeiro (n=827) was conducted investigating the generalisation of SSL methodologies[20]. They found sufficient distributional differences between the datasets but found that the performance of downstream models was not affected by the pretraining cohorts. Previous studies have primarily focused on interpreting results within specific population subsets[17], applying transfer-learning for model pretrained on one cohort to supervised tasks in another cohort[19,20,32], or testing pretrained models on external cohorts[21]. Comparisons between pretraining cohorts have been limited to just two small datasets, both sourced from secondary care settings[14]. However, to date, there has been no comprehensive investigation into how the characteristics of pretraining cohorts influence the quality of learned representations in medical datasets.

A central contribution of our study lies in its novel comparative analysis of CAPE models pretrained on distinct population cohorts. Our findings reveal that along with the cohort size and the pretraining methodology, the health composition and demographic diversity of the training cohort emerge as pivotal determinants for learning robust features. The CAPE models pretrained on secondary care cohorts outperform those pretrained on larger but healthier populations. A plausible explanation for this is that healthier populations tend to exhibit reduced variability, whereas secondary care cohorts encompass a diverse range of cardiac signal patterns. Our contrastive pretraining uses a single pair of ECGs from a patient in each epoch randomly sampled at the training time yet the performance is not solely attributable to the number of patients but also to more ECGs per patient. For example, BIDMC has a smaller number of patients as compared to VUMC but more ECGs per patient, displays a similar performance. This suggests that our random sampling from patient ECGs at the training time can successfully exploit the diversity in ECGs belonging to one patient. When we pretrain our CAPE foundation model[22] on the large composite BCSV cohort, it consistently outperforms models pretrained on individual cohorts (Fig. 2). Therefore, we conclude that the effectiveness of contrastive pretraining depends heavily on the diversity, size, and health characteristics of the underlying population.

**Labeled data distribution affects performance evaluation**

Next, we address the performance evaluation for comparison between pretraining models. In a number of prior works[11], we find that a task-based comparison is undertaken on diverse cohorts. We observe that the same task across diverse cohort is not comparable as the task complexity greatly depends upon the cohort composition. To further illustrate how the performance is affected by cohort distribution, we examine the UKB cohort, which consistently achieves the highest performance across all labels for all CAPE models (Fig. 2). The UKB dataset, stands out due to the narrow interquartile range (IQR) for age and its status as a cohort of healthy volunteers (Table 2). Prior studies suggest that models generally perform better on normal ECGs typical of healthy individuals, and struggle more with learning mappings from abnormal cardiac waveforms[17]. Factors such as the cohort's overall health status, limited age variability, and high data quality likely contribute to the superior model performance observed in this setting. Therefore, performance comparison is only meaningful when evaluated on the same data, ideally using identical train/test splits.

**Combining diverse cohorts for a more effective pretraining of a foundation model with improved generalisation**

We uniquely explore the impact of pretraining data composition on OOD test performance. A comparable work explores domain generalisation for ECG datasets from different sources[19]. The ID and OOD performance for individual ECG classes is evaluated showing a lower performance degradation for their proposed method vs. a baseline supervised model. We used our CAPE model trained on diverse cohorts and then trained an MLP for target tasks on BIDMC only. We then test the MLP on other cohorts to analyse the OOD generalisation. We observe that combining diverse cohorts for contrastive pretraining significantly reduces the ability of downstream models trained on one cohort to generalise to other cohorts (Table 3). We use explainability techniques to understand how the CAPE features map to different cohorts and found clusters related to cohorts (Fig. 4). Further investigation revealed that the mapping was more consistent with device types rather than subject ethnicity (Fig. 4B and Fig. 4C). To address this challenge, we propose a novel in-distribution batch (IDB) strategy that rejects learning spurious technical cohort-specific features but learns more robust features that retain the performance when tested on external cohort. The CAPE model trained with batches randomly formed from different cohorts may encode implicit features like the ECG devices or preprocessing artifacts, that are not clinically meaningful but still help differentiate samples during training. These confounding distributional signals can degrade generalisation to unseen data. In contrast, the CAPE-Z model, avoids overfitting by preserving cohort-level consistency within batches. This promotes the learning of robust and clinically relevant features, resulting in improved OOD performance.

We observe that the IDB demonstrates robust performance for external cohorts (Table 3) but the performance for UKB is still higher than other cohorts. We examine the age distributions across cohorts (Fig. 5A) and the mean absolute error (MAE) for age prediction normalised per age bin (Fig. 5B). The UKB is not only limited to individuals aged 65 to 85 but has a skewed distribution with most subjects between 64 and 66 years, interestingly, this also corresponds to the region with the lowest MAE (Fig. 5B). These findings suggest that the superior performance of UKB is not solely due to the health status of its participants, but also to the over-representation of individuals in the age associated with minimum MAE. This may explain why CODE and SHZS, which also consist of relatively healthy subjects, do not achieve similar performance due to wider age distributions (Fig. 5A).

In this study, we investigate several critical and compelling questions concerning the generalisation capabilities of our CAPE foundational model. Specifically, we examine the role of cohort composition in shaping the robustness of the pretraining process, the performance of downstream supervised models, and their ability to generalise across diverse settings. Building on these insights, we also propose a strategy aimed at enhancing downstream generalisation. We envision that future research can build upon this direction to further advance the understanding of unbiased performance evaluation and the development of more robust foundation models.

**Limitations**

This study aims to investigate how the performance of our contrastive learning methodology is affected by the distributions of the pretraining and label cohorts. While the insights gained are expected to be relevant to other contrastive learning frameworks and supervised training approaches, further experimentation is required to validate these findings in different contexts. In the present work, we use precomputed features without fine-tuning the backbone feature extractor. Future studies should explore how fine-tuning impacts model generalisation and performance. For this study, performance metrics are derived from random splits of each cohort to ensure fair comparisons; as such, the reported results do not reflect performance across the entire cohort. We anticipate that training the prediction head on the full cohort, followed by fine-tuning the model, could further improve performance.

**Conclusions**

This study emphasises the central importance of generalisation in contrastive learning for clinical applications. We find that the ability of a model to perform well across diverse patient cohorts depends critically on the composition of the pretraining data. Specifically, demographic and clinical heterogeneity in the pretraining cohort significantly influences the learned representations and downstream performance. Combining multiple cohorts during contrastive pretraining improves in-distribution performance but reduces out-of-distribution generalisation for our foundation model, when tested on other cohorts. We propose In Distribution (ID batch) method to effectively mitigate this degradation. These findings highlight that careful cohort design and evaluation protocols are essential for building foundation models that generalise reliably across diverse clinical populations, ultimately advancing the development of equitable and robust machine learning systems in healthcare.

## Materials and Methods

### Ethical approvals

All relevant ethical permissions have been obtained for all cohorts explore in the current study. The BIDMC cohort approval is provided by the Beth Israel Deaconess Medical Centre Committee on Clinical Investigations, IRB protocol # 2023P000042. The CODE study is approved by the Research Ethics Committee of the Universidade Federal de Minas Gerais, protocol 49368496317.7.0000.5149. The Institutional Research Board of Zhongshan Hospital (No. 2023-253R) approved the use of SHZS data with a waiver of patient consent. The Vanderbilt component of this study was reviewed and approved by the Institutional Review Board (#212147). The UKB has approval from the Northwest Multi-Centre Research Ethics Committee (application ID 48666). For PTB-XL, the Institutional Ethics Committee approved the publication of the anonymous data in an open-access database (PTB-2020-1).

### Preprocessing

We use leads (I, II, V1-V6), as the remaining four leads (III, aVR, aVF, aVL) do not impart additional information[33]. We apply a bandpass filter (0.5 to 100 Hz) and a notch filter relevant to the mains frequency, re-sample ECGs from different sources to a standard sampling frequency of 400 Hz and use a millivolt scale. The final input to the contrastive learning model is 7-second signal with shape 2800 × 8.

### Pretraining

### Architecture

Fig. 6 presents an overview of the CAPE approach. The ECGs for the same patient are treated as positive views while all other ECGs in the batch are negative views. We use a four-layer ResNet (Residual Network) architecture from prior works[8, 34] as feature generating backbone. The contrastive loss is applied to the non-linear projections of the ECG embeddings (Fig. 6). The 256 features or embeddings learned from the ECGs can be exploited for any downstream supervised training.

**Contrastive Loss**

The contrastive loss employed in this work is the InfoNCE loss[35], applied to nonlinear projections (similar to SimCLR[12]). Given that $z_i z_i z_i$ and $z_i z_j$ are the non-linear projections of representations from two different augmented ECGs belonging to the same patient, the similarity between $z_i$ and $z_j$ is maximized over all other instances in the batch under a softmax function[36], over the similarity scores. The loss function is defined as follows in Equation (1):

$$l_{i,j} = -\log \frac{\exp(\mathrm{sim}(z_i, z_j)/\tau)}{\sum_{k=1}^{2N} I_{[k \neq i]} \exp(\mathrm{sim}(z_i, z_k)/\tau)} \quad (1)$$

Here, $\tau$ is a temperature parameter controlling the sharpness of the distribution, N denotes the number of positive pairs in the batch, and $I_{[k \neq i]} \in \{0,1\}$ is an indicator function that evaluates to 1 when $k \neq i$.

The similarity function $\mathrm{sim}(\cdot,\cdot)$ is defined using cosine similarity in Equation 2:

$$\mathrm{sim}(z_i, z_j) = \frac{z_i^\top z_j}{|z_i| \cdot |z_j|} \quad (2)$$

For batches drawn from a single distribution, we modify the loss as follows. Suppose the training cohort consists of multiple distributions $\{X, Y, ...\}$. Let $z_{iX}$ and $z_{jX}$ be the projections of augmented ECGs from the same patient, both belonging to the distribution X. As each batch contains samples from only one distribution, the similarity between $z_{iX}$ and $z_{jX}$ is maximized over all other instances in the batch that also belong to distribution X. The modified loss is shown in Equation 3:

$$l_{iX,jX} = -\log \frac{\exp(\text{sim}(z_{iX}, z_{jX})/\tau)}{\sum_{k=1}^{2N} I_{[k \neq i]} \exp(\text{sim}(z_{iX}, z_{kX})/\tau)} \quad (3)$$

The total loss over all batches is given in Equation 4:

$$l = \sum_{Z \in \{X,Y,...\}} \sum_{(i,j) \in Z} l_{iZ,jZ} \quad (4)$$

**Training**

The contrastive loss performs best with larger batches due to the larger variation of the negative instances[6], but the computation resources limit the batch size. We use a batch size of 1024 (512 patients) and train the model for 200 epochs using the Adam optimizer[37]. The model is implemented in Tensorflow[38] (2.10.1). The initial learning rate is 0.1 and then decayed according to a half-period cosine schedule[39] (similar to previous approaches[8]). The training time in minutes per epoch on NVIDIA GeForce RTX 3090 is approximately 5 for the BIDMC dataset and 50 for the BCSV dataset.

**Prediction head**

The prediction head uses a multi-layer perceptron (MLP) to learn nonlinear mappings from input features. The MLP architecture is optimized using a grid search over hidden layer sizes of [32, 64, 128, 256] for a two-layer network. For age prediction, the MLP consists of two layers with 256 and 128 neurons, respectively. For sex prediction, both layers contain 256 neurons. The model is trained to predict the target labels using a learning rate of 0.0001, with learning rate decay and early stopping applied to prevent overfitting. The architecture and hyperparameters, including learning rates, were optimized through a limited grid search.

**Performance metrics**

Classification performance for sex prediction is evaluated using the area under the receiver operating characteristic curve (AUC). This metric reflects the ability of the model to distinguish between classes across all possible classification thresholds. The AUC is computed by integrating the true positive rate against the false positive rate over varying thresholds. For age regression, model performance is assessed using the mean absolute error (MAE). MAE measures the average absolute difference between predicted and actual values. To compare CAPE models pretrained on different datasets, statistical significance is assessed using the Kruskal–Wallis test. Comparisons between the In-Distribution Batch (IDB) strategy and training with randomly sampled batches is performed using the Wilcoxon signed-rank test for age prediction (measured by MAE) and the DeLong test for sex classification (measured by AUROC).

**Funding:**

British Heart Foundation (BHF), UK: programme grant funding to FSN and NSP (RG/F/22/110078), clinical research training fellowship to AS, JB and AEM (FS/CRTF/21/24183 and FS/CRTF/24/24624), BHF Centre of Research Excellence funding to FSN, AS and NSP (RE/18/4/34215 and RE/24/130023). Medical Research Council (MRC), UK funding to LP (MR/Y000803/1). AS is also funded by an NIHR Academic Clinical Lectureship. YL is funded by Shanghai Municipal Key Clinical Specialty (shslczdzk01701) and Fudan University AI4S Project (FudanX24AI056). The authors are supported by the National Institute for Health Research Imperial Biomedical Research Centre.

**Acknowledgments:**

The authors would also like to thank the InSIGHT Core in the Center for Healthcare Delivery Science at Beth Israel Deaconess Medical Center for assistance in obtaining primary data.

**Data availability:**

A subset of the CODE dataset, comprising 15% of patients, is publicly available at https://doi.org/10.5281/zenodo.4916206. The PTB-XL dataset is open access and can be obtained from https://physionet.org/content/ptb-xl-plus/1.0.0/. Access to the UK Biobank data is available upon approved application through http://www.ukbiobank.ac.uk/. Due to ethical and legal restrictions, the remaining datasets used in this study are not publicly available.

**Code availability:**

Due to ethical constraints associated with private datasets, the CAPE models used in this study cannot be publicly released. However, to support transparency and reproducibility, we provide a publicly accessible, small-scale working example that demonstrates the key findings of this study. This example utilizes a publicly available subset of the CODE dataset (https://zenodo.org/records/4916206) alongside the PTB-XL dataset, and is available at https://github.com/gulrukhk/CAPE-test/.

We use CAPE features extracted from both datasets in combination with multilayer perceptrons (MLPs), trained on features originally derived from the BIDMC dataset, to perform age and sex prediction. The example includes an external validation scenario, demonstrating performance both with and without the use of the IDB training procedure. Additionally, we include a supervised learning component on the PTB-XL dataset to further demonstrate the effectiveness of the CAPE representations. This includes age regression, sex classification, and diagnostic classification across both superclass and subclass levels.

Together, the provided example and detailed methodological descriptions are sufficient to reproduce the core results and support the main conclusions of this study.

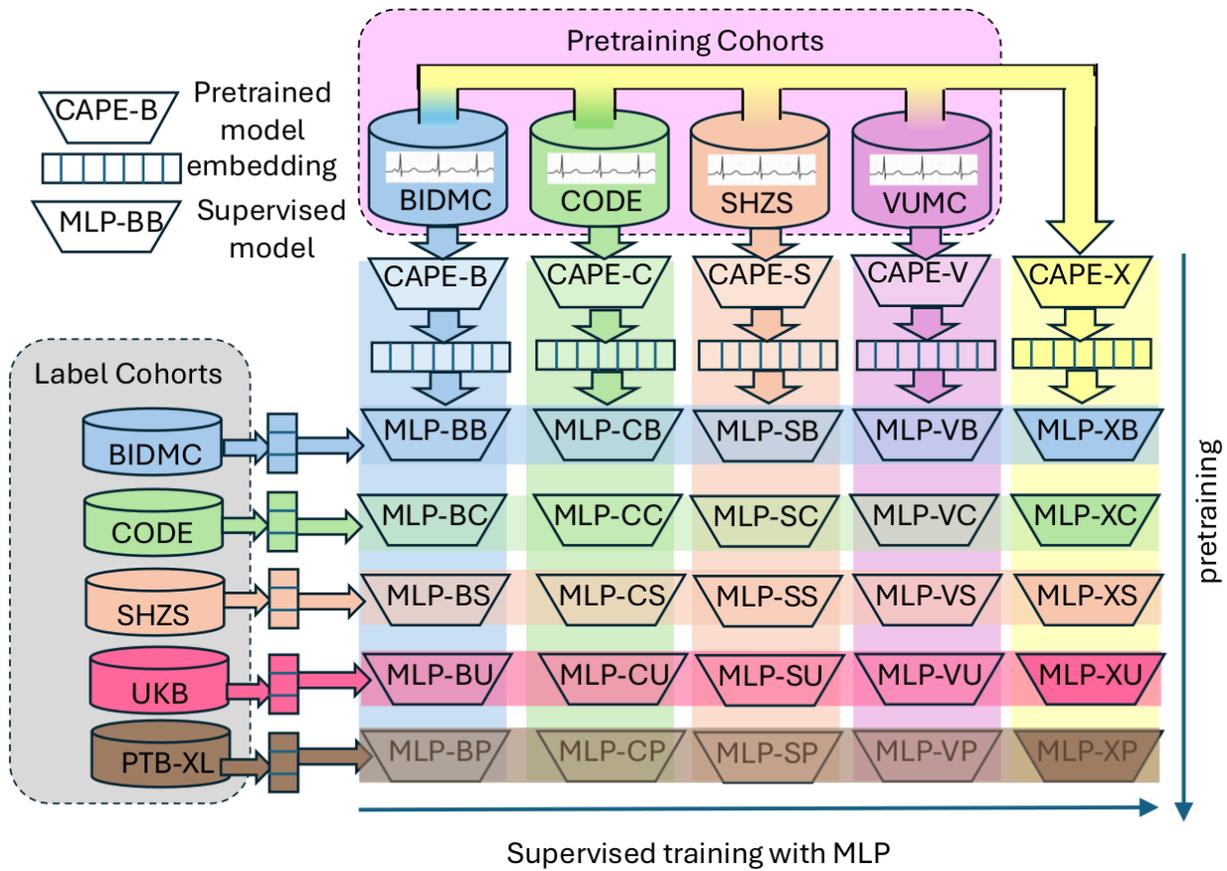

**Fig. 1 Experimental configuration for assessing the impact of cohort health and demographics on model performance.** Five CAPE models are pretrained on distinct datasets: BIDMC, CODE, SHZS, VUMC, and BCSV. For each pretrained model, supervised prediction heads are trained on 10,000 randomly sampled instances and evaluated on 2,000 random test samples, resulting in 25 distinct downstream evaluations. Pretrained models are denoted as CAPE-α, where α refers to the pretraining dataset. The corresponding multilayer perceptron (MLP) models used for

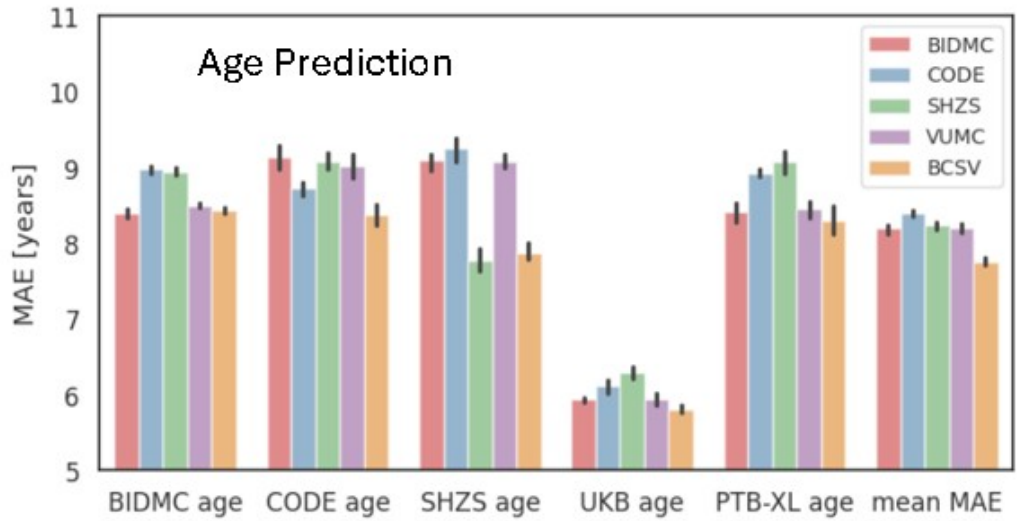

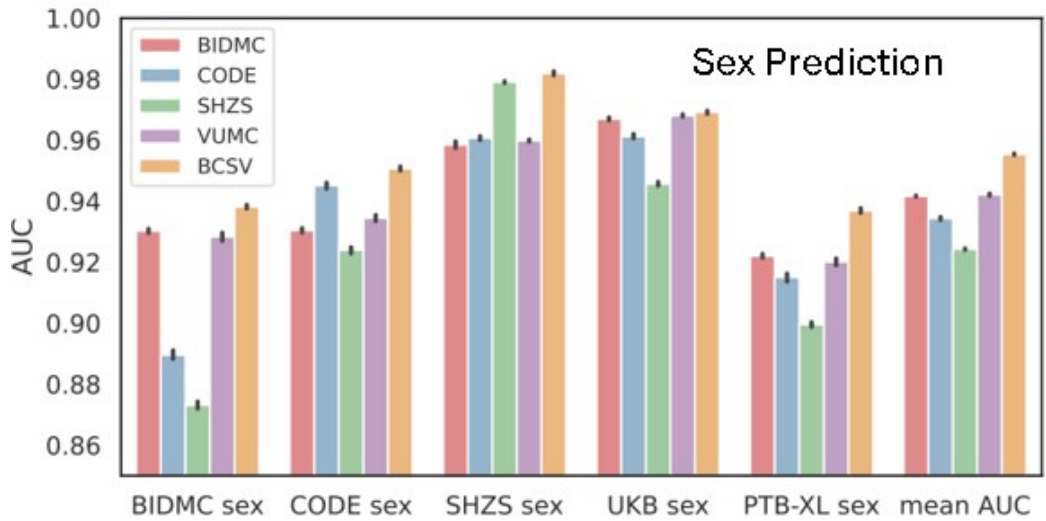

supervised prediction are denoted as MLP-αβ, where α indicates the pretraining dataset and β represents the labeled dataset used for training the prediction head.

**Fig. 2 Performance comparison across different pretraining and labeled datasets.** (A) Age prediction and (B) sex prediction results are shown. Colors correspond to distinct pretraining cohorts. The x-axis represents the labeled dataset used for supervised training, while the y-axis indicates the performance metric under evaluation. Notably, the CAPE model pretrained on BCSV consistently achieves the highest performance across all labeled datasets.

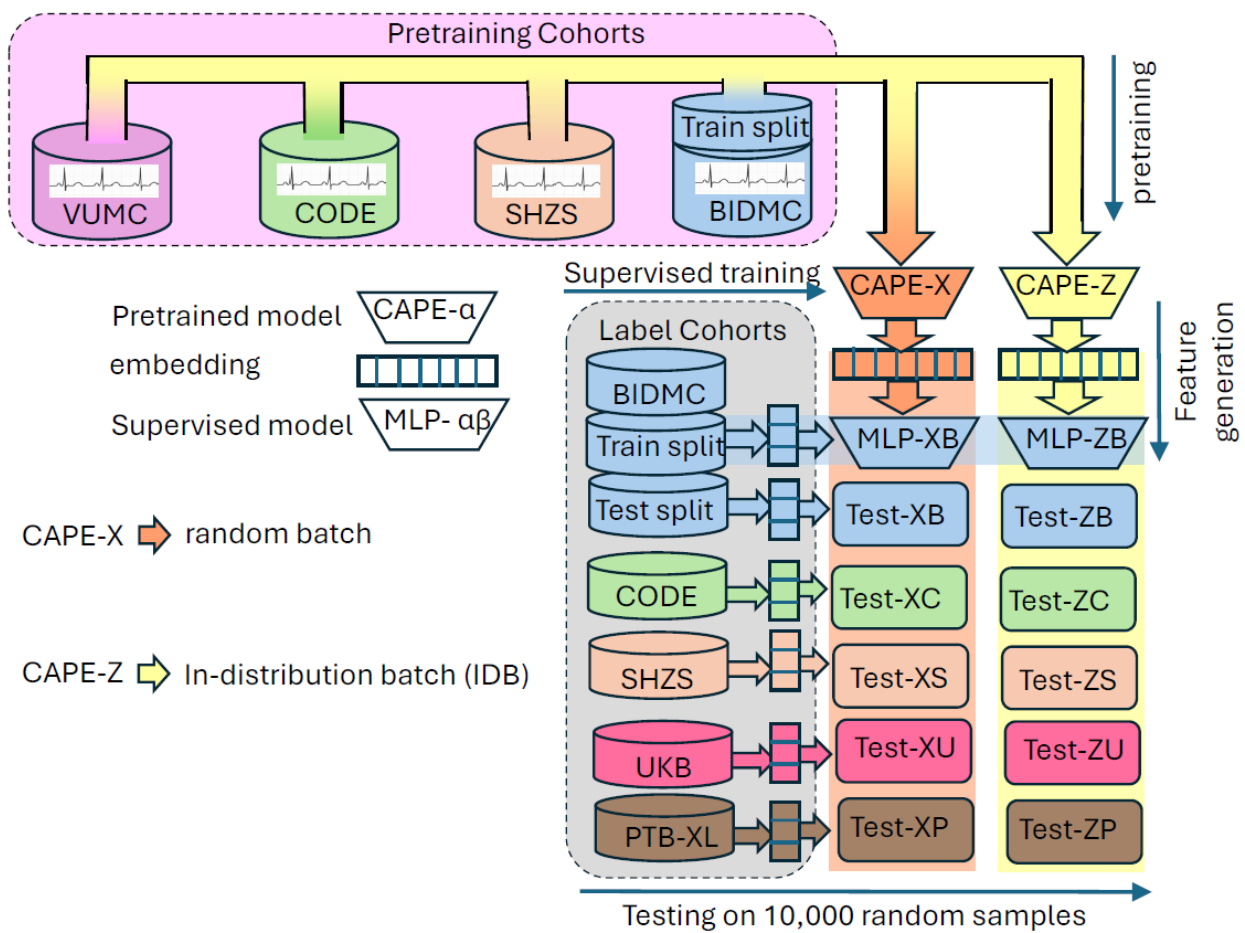

**Fig. 3 Experimental configuration for evaluating out-of-distribution (OOD) performance.** The experiment involves CAPE models pretrained on the combined BCSV dataset, using either random training batches (CAPE-X) or in-distribution (ID) batches (CAPE-Z). Downstream MLP supervised heads are trained on BIDMC labels

across ten independent runs. Models are evaluated on 10,000 random samples drawn from each cohort. Pretrained models are denoted as CAPE-α, where α corresponds to the first letter of the pretraining dataset. The MLP models are denoted as MLP-αβ, with α indicating the pretrained model and β indicating the labeled training cohort. For the test results, Test-αβ denotes evaluation of the model pretrained on α and trained on β.

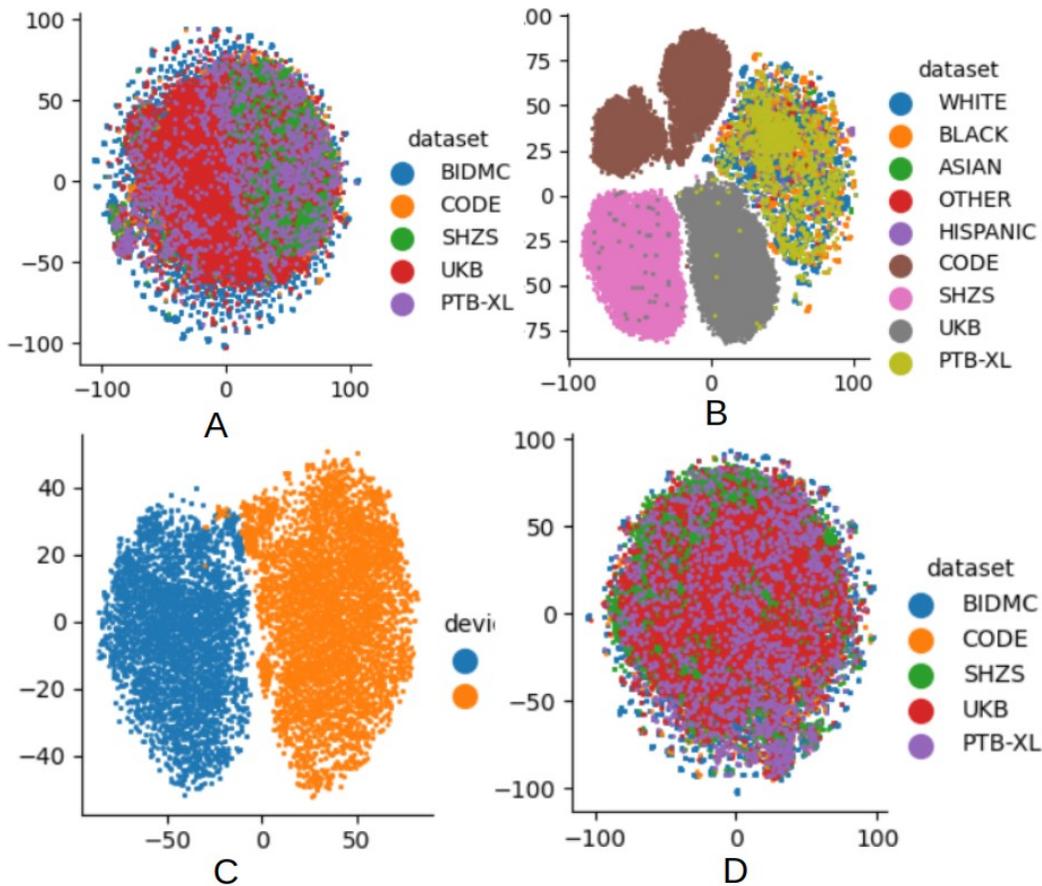

**Fig. 4 t-SNE visualization of learned embeddings for 10,000 random samples from each dataset.** A) Features from a model pretrained on BIDMC (CAPE-B) reveal a continuous gradient mapping across different cohorts. B) Features from a model pretrained on the combined BCSV dataset using random batches (CAPE-X) form distinct clusters for each cohort, except BIDMC and PTB-XL, where different ethnicities within BIDMC cluster together. C) Features from the CODE dataset form two distinct clusters corresponding to the two types of recording devices used. D) Features from a model pretrained on the combined BCSV dataset using in-distribution (ID) batches (CAPE-Z) show no apparent clustering or gradient patterns across cohorts.

**Fig. 5 Impact of age distribution on prediction error for CAPE-Z embeddings with a prediction head trained on BIDMC and tested on external cohorts.** A) Age distributions across different datasets, illustrating mean age and range. B) Mean absolute error (MAE) normalized within each age bin, highlighting the age with minimum

MAE and the overall age range. The distribution of labels significantly influences the aggregated performance metrics, as error varies across the age spectrum. Notably, the UK Biobank dataset exhibits a narrower age range with distinct peaks corresponding to ages associated with the lowest MAE.

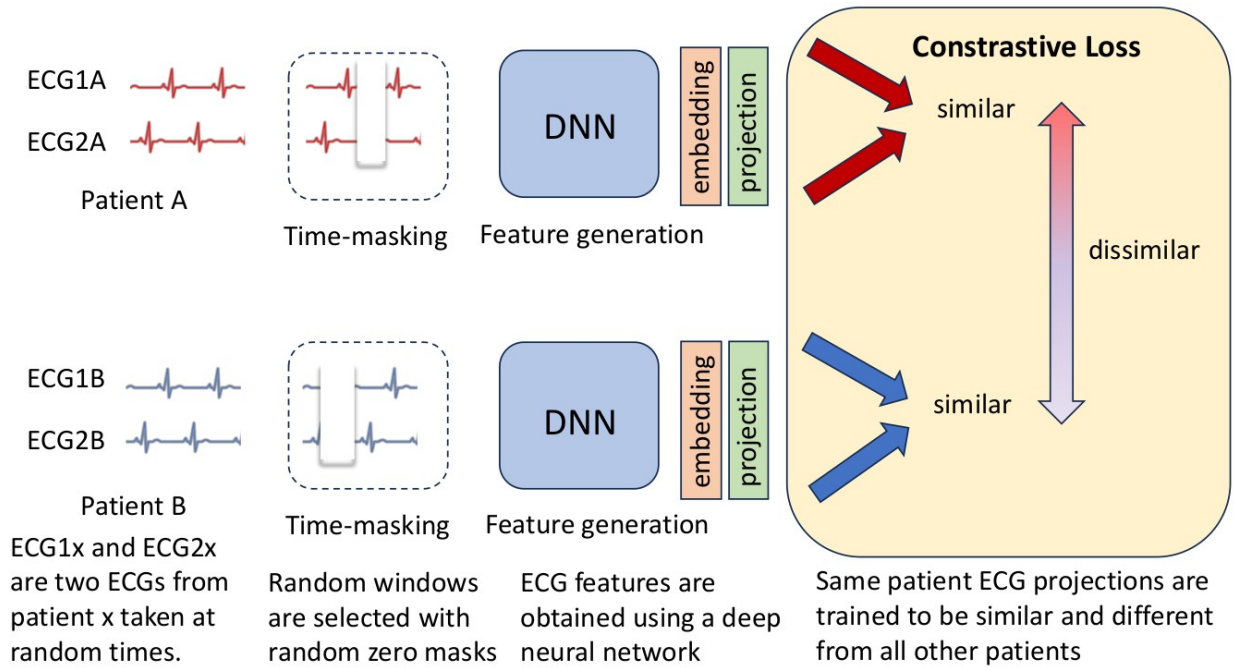

**Fig. 6 Overview of the CAPE pretraining framework.** Patient A has two ECG recordings (ECG1A and ECG2A), and Patient B, another patient in the same batch, has ECG1B and ECG2B. Each ECG undergoes temporal augmentations before being encoded by the network into a 256-dimensional feature vector. The contrastive loss encourages embeddings from the same patient's ECGs to be closer together while pushing apart embeddings from different patients within the batch.

**Table 1 Baseline characteristics of the pretraining cohorts included in this study.**

|  | BIDMC∗ | CODE∗ | SHZS∗ | VUMC∗ |
|---|---|---|---|---|
| Location | US | Brazil | China | US |
| Care level | secondary | primary | Primary | Secondary |
| Patients∗ | 127,041 | 424,577 | 420,956 | 252,306 |
| ECGs | 1,106,886 | 1,123,903 | 1,560,551 | 1,412,012 |
| Age mean | 57.99 | 56.00 | 52.08 | 58.05 |
| Age IQR | 23.02 | 23.00 | 27.00 | 24.30 |
| Male | 63,006 (50%) | 165,285 (39%) | 233,808 (56%) | 127,898 (51%) |
| Female | 64,035 (50%) | 259,292 (61%) | 187,148 (44%) | 121,008 (49%) |
| Hispanic | 7,077 | — | — | — |
| White | 84,265 | — | — | — |
| Black | 17,778 | — | — | — |
| Asian | 5,315 | — | — | — |
| Other | 12,606 | — | — | — |

∗ Pretraining cohort with patients with more than one ECG

**Table 2** Baseline characteristics of the cohorts explored for the downstream supervised tasks (BIDMC, CODE, SHZS, UKB, and PTB-XL).

|  | BIDMC | CODE | SHZS | UKB | PTB-XL |
|---|---|---|---|---|---|
| Location | US | Brazil | China | UK | Germany |
| Care level | secondary | primary | primary | primary | Secondary |
| Patients* | 189,542 | 46,986 | 43,209 | 66,402 | 18,869 |
| ECGs | 1,169,387 | 50,000 | 50,000 | 70,655 | 21,799 |
| Age mean | 55.62 | 53.35 | 54.59 | 65.35 | 62.36 |
| Age IQR | 26.37 | 25.00 | 21.00 | 12.00 | 23.00 |
| Male | 90,793 (48%) | 18,546 (39%) | 23,171 (54%) | 32,191 (49%) | 9,640 (51%) |
| Female | 98,749 (52%) | 28,440 (61%) | 20,037 (46%) | 34,211 (51%) | 9,229 (49%) |
| Hispanic | 10,248 | – | – | – | – |
| White | 123,063 | – | – | – | – |
| Black | 24,251 | – | – | – | – |
| Asian | 8,924 | – | – | – | – |
| Other | 23,056 | – | – | – | – |

**Table 3 | Performance comparison of foundation models pretrained using random batch versus in-distribution batch (IDB) strategies. The prediction head is trained on BIDMC labels across ten independent runs, with evaluation conducted on 10,000 random test samples from external cohorts. Performance metrics are reported as the mean with standard error (SE) in parentheses: e.g. mean absolute error (MAE) with mean of 7.880 and standard error 0.0028 is presented as 7.880(028), and the area under the curve (AUC) of mean 0.939 and standard error 0.0006 as 0.939(06).**

| No. | Model | BIDMC | CODE | SHZS | UKB | PTB-XL | mean |
|---|---|---|---|---|---|---|---|
| | | Age prediction in years (MAE) | | | | | |
| 1. | CAPE-X | **7.880(028)** | 24.232(∗) | 14.326(∗) | 5.786(018) | 7.786(019) | 12.00 |
| 2. | CAPE-Z | 7.833(031) | **7.962(031)** | **7.378(020)** | **5.875(017)** | **7.681(017)** | **7.346** |
| 3. | p-value[!] | 0.004 | 0.002 | 0.002 | 0.002 | 0.010 | - |
| | | Sex prediction (AUC) | | | | | |
| 4. | CAPE-X | **0.939(06)** | 0.600(∗) | 0.659(∗) | 0.974(04) | 0.937(05) | 0.822 |
| 5. | CAPE-Z | 0.937(08) | **0.935(08)** | **0.942(07)** | **0.977(03)** | **0.947(06)** | **0.948** |
| 6. | p-value[$] | <0.0001 | <0.0001 | <0.0001 | <0.0001 | <0.0001 | - |

∗ Out of range with SE(MAE) ≥ 1.0 or SE(AUC) ≥ 0.01.
[!] Wilcoxon signed-rank test
[$] DeLong test

**Contributions**